\documentclass[11pt]{article}
\usepackage{coling2018}
\usepackage{times}
\usepackage{url}
\usepackage{latexsym}
\usepackage{amsmath,amssymb,array,subfig,graphicx,caption}
\usepackage{multirow}

\title{An Interpretable Reasoning Network for Multi-Relation Question Answering}

\author{
  Mantong Zhou \quad Minlie Huang\thanks{*Corresponding author: Minlie Huang (aihuang@tsinghua.edu.cn).} \quad Xiaoyan Zhu\\
   State Key Lab. of Intelligent Technology and Systems,\\ National Lab. for Information Science and Technology, \\
  Dept. of Computer Science and Technology, Tsinghua University, Beijing, PR China \\
  {\tt zmt.keke@gmail.com}, \quad
  {\tt aihuang@tsinghua.edu.cn}, \\
  {\tt zxy-dcs@tsinghua.edu.cn}\\
  }

\date{}

\newcolumntype{I}{!{\vrule width 0pt}}

\begin{document}
\maketitle
\begin{abstract}
  Multi-relation Question Answering is a challenging task, due to the requirement of elaborated analysis on questions and reasoning over multiple fact triples in knowledge base. In this paper, we present a novel model called {\bf Interpretable Reasoning Network} that employs an interpretable, hop-by-hop reasoning process for question answering. The model dynamically decides which part of an input question should be analyzed at each hop; predicts a relation that corresponds to the current parsed results; utilizes the predicted relation to update the question representation and the state of the reasoning process; and then drives the next-hop reasoning. Experiments show that our model yields state-of-the-art results on two datasets. More interestingly, the model can offer traceable and observable intermediate predictions for reasoning analysis and failure diagnosis, thereby allowing manual manipulation in predicting the final answer.
\end{abstract}

\section{Introduction}
\label{intro}

%
%

\blfootnote{
    %
    %
    %
    %
     \hspace{-0.65cm}  
     This work is licenced under a Creative Commons 
     Attribution 4.0 International Licence.
     Licence details:
     \url{http://creativecommons.org/licenses/by/4.0/}
    %
    %
}

Open-domain Question Answering (QA) has always been a hot topic in AI and this task has recently been facilitated by large-scale Knowledge Bases~(KBs) such as Freebase~\cite{Bollacker2008Freebase}. However, due to the variety and complexity of language and knowledge, open-domain question answering over knowledge bases (KBQA) is still a challenging task.

Question answering over knowledge bases falls into two types, namely single-relation QA and multi-relation QA, as argued by Yin et al.~\shortcite{QACNN}. Single-relation questions, such as { \em ``How old is Obama?"}, can be answered by finding one fact triple in KB, and this task has been widely studied~\cite{SimpleQuestion,xu2016enhancing,EVINET}. In comparison, reasoning over multiple fact triples is required to answer multi-relation questions such as {\em ``Name a soccer player who plays at forward position at the club Borussia Dortmund."} 
where more than one entity and relation are mentioned. 
Compared to single-relation QA, multi-relation QA is yet to be addressed.

Previous studies on QA over knowledge bases can be roughly categorized into two lines: semantic parsing and embedding-based models. Semantic parsing models ~\cite{yih2014semantic,Yih2016parse} obtain competitive performance at the cost of hand-crafted features and manual annotations, but lack the ability to generalize to other domains. In contrast, embedding-based models~\cite{EmbedQA,qawithcrossattention,D17qtype} can be trained end-to-end with weak supervision, but existing methods are not adequate to handle multi-relation QA due to the lack of reasoning ability. 

Recent reasoning models~\cite{KVMemN2N,P17-1018} mainly concentrate on Reading Comprehension (RC) which requires to answer questions according to a given document. However, transferring existing RC methods to KBQA is not trivial. For one reason, the focus of reasoning in RC is usually on understanding the document rather than parsing questions.
For another reason, existing reasoning networks are usually designed in a black-box style, making the models less interpretable. While in multi-relation question answering, we believe that an interpretable reasoning process is essential.

In this paper, we propose a novel Interpretable Reasoning Network (IRN) to equip QA systems with the reasoning ability to answer multi-relation questions. 
Our central idea is to design an interpretable reasoning process for a complex question: the reasoning module decides which part of an input question should be analyzed at each hop, and predicted a KB relation that corresponds to the current parsed results. The predicted relation will be used to update the question representation as well as the state of the reasoning module, and helps the model to make the next-hop reasoning. 
At each hop, an entity will be predicted based on the current state of the reasoning module.


Different from previous models, our model is {\em \textbf{interpretable}} in that the predicted relation and entity at each hop are {\em \textbf{traceable and observable}}. At each hop our model has a specific aim to find an appropriate relation based on the iterative analysis of a question, and intermediate output at each hop can be interpreted by the corresponding linked entity. In this manner, IRN offers the ability of visualizing {\em \textbf{a complete reasoning path}} for a complex question, which facilitates reasoning analysis and failure diagnosis, thereby allowing manual manipulation in answer prediction as detailed in our experiments.

The contributions of this paper are in two folds:
\begin{enumerate} 
	\item We design an Interpretable Reasoning Network which can make reasoning on multi-relation questions with multiple triples in KB. Results show that our model obtains state-of-the-art performance.

	\item Our model is more interpretable than existing reasoning networks in that the intermediate entities and relations predicted by the hop-by-hop reasoning process construct traceable reasoning paths to clearly reveal how the answer is derived.

\end{enumerate}

\section{Related Works}
\label{sect:related works}

Recent works on QA can be roughly classified into two types: one is semantic-parsing-based and the other is embedding-based.
Semantic parsing approaches map questions to logical form queries~\cite{Pasupat2015Compositional,Yih2016parse,D17interpretableQA}. These systems are effective but at the cost of heavy data annotation and pattern/grammar engineering.
What's more, parsing systems are often constrained on a specific domain and broken down when executing logical queries on incomplete KBs.

Our work follows the line of Embedding-based models~\cite{EmbedQA,multicolum,xu2016enhancing,qawithcrossattention,D17qtype} which are recently introduced into the QA community where questions and KB entities are represented by distributed vectors, and QA is formulated as a problem of matching between vectors of questions and answer entities. These models need less grammars as well as annotated data, and are more flexible to deal with incomplete KBs. To make better matching, subgraphs of an entity in KB~\cite{SubgraphEmbed}, answer aspects~\cite{multicolum,qawithcrossattention} and external contexts~\cite{xu2016enhancing} can be used to enrich the representation of an answer entity. Though these methods are successful to handle simple questions, answering multi-relation questions or other complex questions is far from solved, since such a task requires reasoning or other elaborated processes.

Our work is also related to recent reasoning models which focus on Reading Comprehension where memory modules are designed to comprehend documents. State-of-the-art memory-based Reading Comprehension models~\cite{MemN2N,DynamicMem,Reasonet,P17-1018,Celikyilmaz2017Scaffolding} make interactions between a question and the corresponding document in a multi-hop manner during reasoning. MemNN~\cite{MemNN}, KVMemN2N~\cite{KVMemN2N} and EviNet~\cite{EVINET} transferred the reading comprehension framework to QA where a set of triples is treated as a document and a similar reasoning process can be applied. However, reading comprehension makes reasoning over documents instead of parsing the questions. 

Other studies applying hop-by-hop inference into QA can be seen in Neural Programmer~\cite{Neelakantan2015NeuralProgrammer,neelakantan2016neuralprogrammer} and Neural Enquirer~\cite{NeuralEnquirer}, where deep networks are proposed to parse a question and execute a query on tables. 
However, Neural Programmer needs to predefine symbolic operations, while Neural Enquirer lacks explicit interpretation. Mou et al.~\shortcite{Mou2016Coupling} proposed a model coupling distributed and symbolic execution with REINFORCE algorithm, however, training such a model is challenging.
Neural Module Network~\cite{NMN,composeQA}  customized network architectures for different patterns of reasoning, making the reasoning network interpretable. However, a dependency parser and the REINFORCE algorithm are required.

\section{Interpretable Reasoning Network}
\label{sect:overview}

\subsection{Task Definition}
\label{sect:task}
Our goal is to offer an interpretable reasoning network to answer multi-relation questions. Given a question $q$ and its topic entity or subject $e_s$ which can be annotated by some NER tools, the task is to find an entity $a$ in KB as the answer.

In this work, we consider two typical categories of multi-relation questions, a path question~\cite{Traversing} and a conjunctive question~\cite{WC2014}, while the former is our major focus.

\noindent{\bf A path question} contains only one topic entity (subject $e_s$) and its answer (object $a$) can be found by walking down an answer path consisting of a few relations and the corresponding intermediate entities.  We define an {\bf answer path} as a sequence of entities and relations in KB which starts from the subject and ends with the answer like { $e_s\xrightarrow{r_1} e_1\xrightarrow {r_2} ...\xrightarrow{r_n} a$}. Relations~($r_i$) are observable (in various natural language forms) in the question, however, the intermediate entities ($e_1\cdots e_H$) are not. For example, for question {\it``How old is Obama's daughter?"}, the subject is {\em Barack\_Obama} and the answer path is {\small \em Barack\_Obama$\xrightarrow{Children}$Malia\_Obama$\xrightarrow{Age}$18}. 
Note that since there are 1-to-many relations\footnote{For instance, relation \textit{Children} is one-to-many, where a person may have more than one child.}, the range of the intermediate entities can be large, resulting in more than one answer path for a question. 


\noindent{\bf A conjunctive question} is a question that contains more than one subject entity and the answer can be obtained by the intersection of results from multiple path queries. For instance, the question {\em ``Name a soccer player who plays at forward position at the club Borussia Dortmund."} has a possible answer as the intersection of results from two path queries\footnote{Superscript -1 stands for the inverse relation.} {\small \em $FORWARD\xrightarrow {plays\_position^{-1}}Marco\_Reus$} and {\small \em $Borussia\_ Dortmund\xrightarrow {plays\_in\_club^{-1}} Marco\_Reus$}. The details for dealing with conjunctive questions are shown in Fig~\ref{fig:frame}.

\subsection{Overview}
The reasoning network has three modules: input module, reasoning module, and answer module. The input module encodes the question into a distributed representation and updates the representation hop-by-hop according to the inference results of the reasoning module. The reasoning module initializes its state by the topic entity of a question and predicts a relation on which the model should focus at the current hop, conditioned on the present question and reasoning state. The predicted relation is utilized to update the state vector and the question representation hop-by-hop. The answer module predicts an entity conditioned on the state of the reasoning module.

\begin{figure*}[htbp]
	\centering
	\small
	\includegraphics[width=0.78\linewidth]{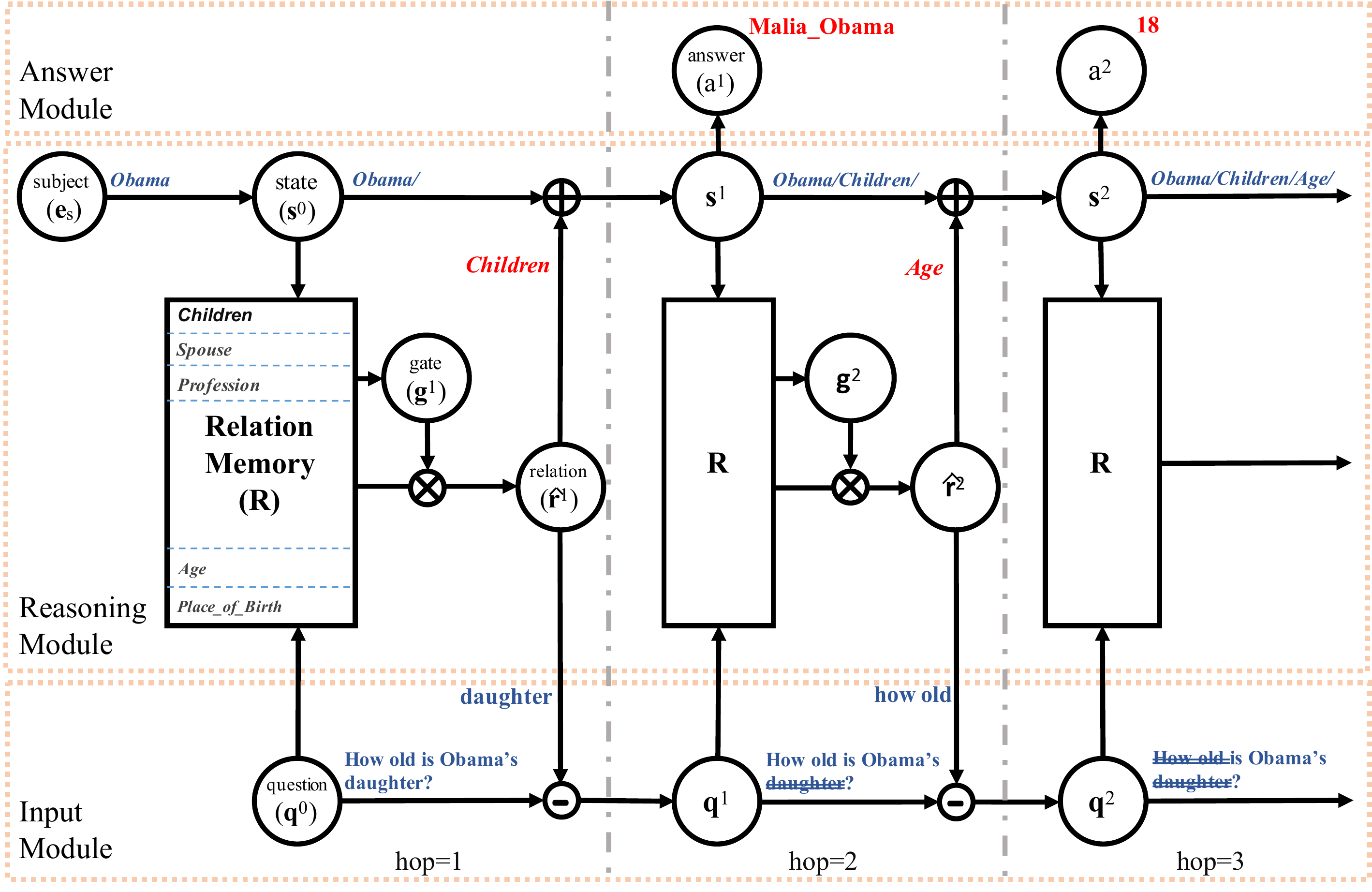}
	\caption{\small Interpretable Reasoning Network. At each hop IRN computes the probability of selecting the next relation as $\boldsymbol{g}^h$ and obtains a predicted relation $\boldsymbol{\hat{r}}^h$. The predicted relation $\boldsymbol{\hat{r}}^h$ is used to update the question $\boldsymbol{q}^h$ and the state $\boldsymbol{s}^{h}$ with different projections. The state is initialized by subject as $\boldsymbol{s}^{0}=\boldsymbol{e}_s$. The answer path ($Obama\xrightarrow{Children}Malia\_Obama\xrightarrow{Age}18$) is composed of the predicted relations and entities (in red).}
	\label{fig:model}
\end{figure*}

The process can be illustrated by the example as shown in Figure~\ref{fig:model}. For question {\em ``How old is Obama's daughter?"}, the subject entity { \em Barack\_Obama} is utilized to initialize the state vector. IRN predicts the first relation { \em ``Children"} at the first hop. The { \em ``Children"} relation is added to the state vector to encode the updated parsing result, and the corresponding natural language form of this relation in the question (here is { \em ``daughter"}) is subtracted from the question to avoid repeatedly analyzing the relation-relevant word { \em ``daughter"}. This procedure is repeated until the { \em Terminal} relation is predicted.

\subsection{Input Module}
The input module encodes a question to a vector representation and updates the representation of the question at each hop of the reasoning process: the predicted relation will be subtracted from the current representation to compel the reasoning process to attend to other words that should be analyzed.

Formally, the question $\boldsymbol{X}=x_1,x_2,...,x_n$ can be initialized by the sum of the word embeddings and updated by subtracting the relation predicted by the reasoning module at the previous hop:
\begin{eqnarray}
\boldsymbol{q}^0&=&\sum_{i=1}^n \boldsymbol{x}_i\\
\boldsymbol{q}^h&=&\boldsymbol{q}^{h-1} - \boldsymbol{M}_{rq}\boldsymbol{\hat{r}}^{h} \label{eq:Mrq}
\end{eqnarray}
where $\boldsymbol{M}_{rq}$ is a matrix projecting the KB relation space to the natural language question space, $\boldsymbol{q}^{h-1}$ is the question representation at hop $h-1$, and $\boldsymbol{\hat{r}}^{h}$ defined by Eq.~\ref{eq:r_hat} is the predicted relation at hop $h$. The intuition of such update is that the already analysed part of the question should not be parsed again.

Representing a question as a bag of words might be too simple. However, this method works well in our setting.
Future work would consider other sophisticated encoders such as CNN or LSTM.

\subsection{Reasoning Module}
\label{sec:reasoning}
The reasoning module aims to 
attend to a particular part of the question at each hop, predict an associated relation in knowledge base, and then update its state. 

The reasoning module takes as input the previous state vector ($\boldsymbol{s}^{h-1}$) and the previous question vector ($\boldsymbol{q}^{h-1}$), and then predicts a relation ($\boldsymbol{\hat{r}}^h$) based on the analysis at the current hop. Once the predicted relation ($\boldsymbol{\hat{r}}^h$) is obtained, the relation will be used to update the next question representation ($\boldsymbol{q}^{h}$) and the state of the reasoning module ($\boldsymbol{s}^{h}$). In this manner, the reasoning network is traceable and interpretable. 

The process can be formally described by the following equations\footnote{$\boldsymbol{a}^T\boldsymbol{b}$ is the inner-product of vector $\boldsymbol{a}$ and $\boldsymbol{b}$.}:
\begin{align}
	g_j^h = P(r^h=r_j|\boldsymbol{q}^{h-1},\boldsymbol{s}^{h-1})&=\operatorname{softmax}((\boldsymbol{M}_{rq}\boldsymbol{r}_j)^T\boldsymbol{q}^{h-1} + (\boldsymbol{M}_{rs}\boldsymbol{r}_j)^T\boldsymbol{s}^{h-1} )\label{eq:g}\\
	\boldsymbol{\hat{r}}^h &= \sum_j g_j^h * \boldsymbol{r}_j \label{eq:r_hat} \\
	\boldsymbol{s}^h &= \boldsymbol{s}^{h-1} + \boldsymbol{M}_{rs}\boldsymbol{\hat{r}}^h \label{eq:state}
\end{align}
where $\boldsymbol{r}_j$ is the embedding of a relation in KB and all the relation embeddings are stored in a static memory $R$, and $\boldsymbol{s}^h$ is the state of the reasoning module at hop $h$.
$g_j^h$ is the probability of selecting the $j^{th}$ relation in KB and $\boldsymbol{M}_{rs}$ is the projection matrix mapping $\boldsymbol{r}$ from the relation space to the state space. $\boldsymbol{M}_{rq}$ is the same projection matrix used in Eq.~\ref{eq:Mrq} to map $\boldsymbol{r}$ from the relation space to the question space.

We initialize the state vector with the topic entity (subject) $\boldsymbol{s}^0 = \boldsymbol{e}_s$. IRN will learn to enrich the state representation hop by hop, for instance, at the first hop $\boldsymbol{s}^1 \approx \boldsymbol{e}_s+\boldsymbol{r}_1$, and at the second hop $\boldsymbol{s}^2 \approx \boldsymbol{e}_s+\boldsymbol{r}_1+\boldsymbol{r}_2$, intuitively. In this manner, the state vector encodes historical information. 

In order to signify when the reasoning process should stop, we augment the relation set with the {\em Terminal} relation. Once the reasoning module predicts the { \em Terminal} relation, the reasoning process will stop, and the final answer will be the output when the last non-terminal relation is added to the state $\boldsymbol{s}$.

\subsection{Answer Module}
\label{sec:ans}
The answer module chooses the corresponding entity from KB at each hop (denoted as $a^h$). At the last hop, the selected entity is chosen as the final answer, while at the intermediate hops, the predictions of these entities can be inspected to help reasoning analysis and failure diagnosis.

More formally, an entity at each hop can be predicted as follows:
\begin{align}
	\boldsymbol{e}^h&=\boldsymbol{M}_{se}\boldsymbol{s}^h \label{eg:Mse}\\
	o^h_j&=P(a^h=e_j|\boldsymbol{s}^h)=\operatorname{softmax}(\boldsymbol{e}_j^T\boldsymbol{e}^h) \label{eq:o}  
\end{align}

$\boldsymbol{M}_{se}$ is used to transfer from the state space ($\boldsymbol{s}^h$) to the entity space ($\boldsymbol{e}^h$) to bridge the representation gap between the two spaces. $\boldsymbol{e}_j$ is the embedding vector of the $j^{th}$ entity in KB.

\subsection{Loss Function}
\label{sec:loss}

We adopt cross entropy to define the loss function. 
The first loss term is defined on the intermediate prediction of relations, while the second term on the prediction of entities.

The loss on one instance is defined as follows:
\begin{align}
	\mathcal{L}=\sum_h \mathcal{C}_r(h) &+\lambda \mathcal{C}_a(h)\label{eq:loss}\\
	\mathcal{C}_r(h) = -\sum_{j=1}^{n_r}[\hat{g}_j^h \ln g_j^h] \ &, \  \mathcal{C}_a(h) = -\sum_{i=1}^{n_e}[ \hat{o}^h_i\ln o^h_i] \notag
\end{align}
where $n_r$/$n_e$ is the number of relations/entities in KB respectively, $\boldsymbol{\hat{g}}^h$ is the gold distribution (one-hot) over relations at hop $h$, $\boldsymbol{g}^h$ is the predicted distribution defined by Eq.~\ref{eq:g}, $\boldsymbol{\hat{o}}$ is the gold distribution over entities, which is also one-hot representation, and $\boldsymbol{o}$ is defined by Eq.~\ref{eq:o}. $\lambda$ is a hyper parameter to balance the two terms.

Note that the training data is in the form of ($q,<e_s,r_1,e_1,...,a>$), which indicates that the model can incorporate supervision not only from the final answer (referred to as IRN-weak), but also from the intermediate relations and entities along the answer path (referred to as IRN).

\subsection{Multitask Training for KB Representation}
\label{sec:multi-training}
In order to incorporate more constraints from KB\footnote{Constraint from KB means that two entities and a relation form a triple in KB, as $(e_s, r, e_o)$.}, we learn the embeddings of entities and relations as well as the space transition matrix with a multitask training schema. For a given fact triple in KB, $(e_s,r,e_o)$, the representations of the entities and the relation apply the following constraint:
\begin{equation}
\boldsymbol{M}_{se}(\boldsymbol{e}_s + \boldsymbol{r}) = \boldsymbol{e}_o\label{eq:multi-task}
\end{equation}
where $\boldsymbol{e}_s,r,\boldsymbol{e}_o$ are embeddings of the subject (or head) entity, relation, and the object (or tail) entity. This idea is inspired by TransE~\cite{TransE}, but we adopt $\boldsymbol{M}_{se}$ (see Eq.~\ref{eg:Mse}) as a transfer matrix to bridge the representation gap between the state space (here $\boldsymbol{e}_s+\boldsymbol{r}=\boldsymbol{s}$) and the entity space (here $\boldsymbol{e}_o=\boldsymbol{e}$). 

The parameters are updated with a multi-task training schema. We first learn the KB embeddings $\boldsymbol{e}$/$\boldsymbol{r}$ and the transformation matrix $\boldsymbol{M}_{se}$ to fit Eq.~\ref{eq:multi-task} with several epoches. This is the task of KB embedding training. Then, we update all the parameters of IRN under supervision from the QA task with one epoch, which is the task of QA training.
We run the two tasks iteratively.

With the help of the auxiliary KB embedding training, IRN not only utilizes the additional information from KB to make better inferences, but also has an ability to deal with incomplete answer paths. For example, even if the connection between { \em Barack\_Obama} and { \em Malia\_Obama} is not present in KB, our model can still make correct prediction thanks to {$\boldsymbol{M}_{se}(\boldsymbol{e}_{Barack\_Obama}+\boldsymbol{r}_{Children})\approx \boldsymbol{e}_{Malia\_Obama}$}. 

\subsection{Dealing with Conjunctive Questions}
\label{sect:conj-query}
IRN is not limited to only path questions. For a conjunctive question that contains more than one topic entity, the answer can be found by executing multiple IRNs with the same parameters in parallel and then obtaining the intersection of individual results.
\begin{figure}[htbp]
	\centering
	\small
	\includegraphics[height=0.25\linewidth]{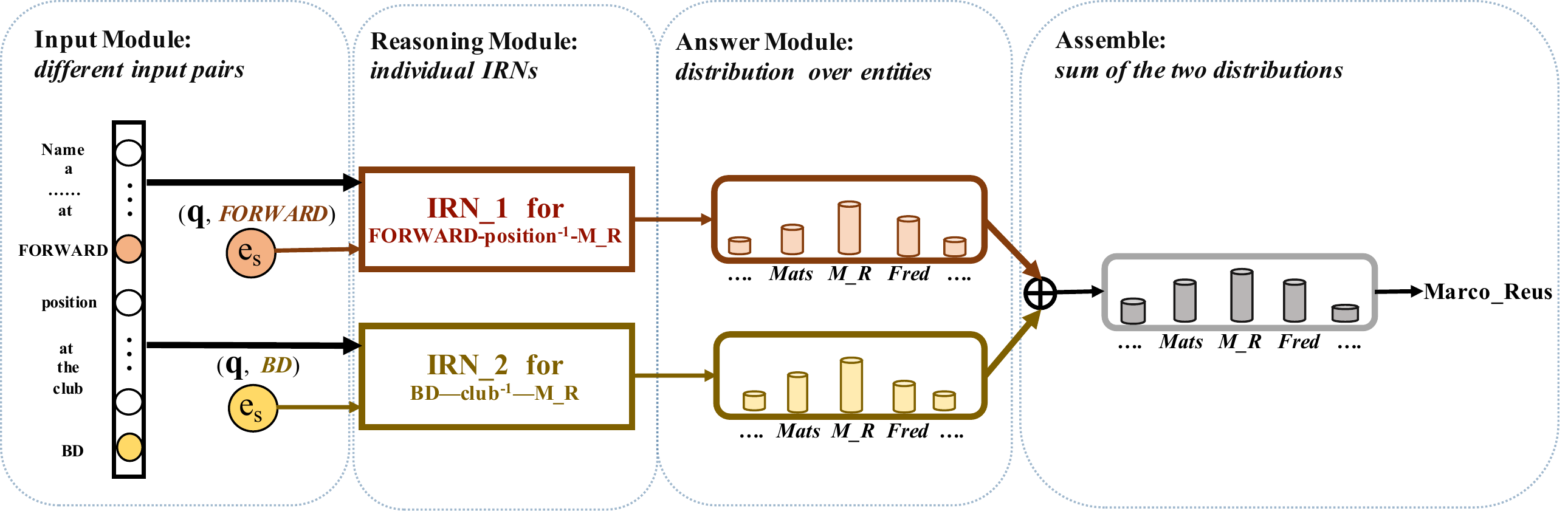}
	\caption{\small An assembly of two IRNs to handle a conjunctive question with two subjects. Different IRNs take as input the same question but different subjects and output the distribution over the candidate answers. The final answer is selected after summing the two distributions.
	}
	\label{fig:frame}
\end{figure}

This process is exemplified by Figure~\ref{fig:frame}. The input question {\em ``Name a soccer player who plays at forward position at the club Borussia Dortmund"} has two subject entities, {\em ``FORWARD"} and {\em ``Borussia\_Dortmund(BD)"}. One IRN (IRN\_1) takes the original question and {\em ``FORWARD"} as input, and then predicts possible objects for path query {\small \em ``FORWARD$\xrightarrow{plays\_position^{-1}}$ ?(Marco\_Reus)"}.\footnote{The question mark indicates the entity to be predicted and the entity in bracket is the expected answer.}
The output is a distribution over entities.
Similarly, another IRN (IRN\_2) tackles the path query {\small \em ``BD$\xrightarrow{plays\_in\_club^{-1}}$?(Marco\_Reus)"} where the input is the same question but another subject entity {\em ``Borussia Dortmund(BD)"}. After summing the two output distributions, the answer {\em ``Marco\_Reus"} is chosen  with the largest probability.

\section{Data Preparation}
\label{sect:datasets}
We prepared two KBQA datasets to evaluate our Interpretable Reasoning Network: one is PathQuestion\footnote{This dataset is available at \url{https://github.com/zmtkeke/IRN}}, constructed by ourselves, and the other is WorldCup2014, adopted from ~\cite{WC2014}.

\subsection{PathQuestion}
We adopted two subsets of  Freebase~\cite{Bollacker2008Freebase} as Knowledge Bases to construct the PathQuestion (PQ) and the PathQuestion-Large (PQL) datasets.
We extracted paths between two entities which span two hops ($e_s\xrightarrow{r_1}e_1\xrightarrow{r_2}a$, denoted by -2H) or three hops ($e_s\xrightarrow{r_1}e_1\xrightarrow{r_2}e_2\xrightarrow{r_3}a$, denoted by -3H)  and then generated natural language questions with templates.  
To make the generated questions analogical to real-world questions, we included paraphrasing templates and synonyms for relations by searching the Internet and two real-world datasets,
WebQuestions~\cite{WebQuestions} and WikiAnswers~\cite{Fader2013Paraphrase}. In this way, the syntactic structure and surface wording of the generated questions have been greatly enriched. 

PQL is more challenging than PQ in that PQL utilizes larger KB and provides less training instances. The statistics are shown in Table~\ref{tab:dataset} and more details are described in the Appendix~\ref{sec:supplemental}.

\subsection{WorldCup2014}
We also evaluated our model on the WorldCup2014 (WC2014) dataset constructed by~\cite{WC2014}. The dataset contains single-relation questions (denoted by WC-1H), two-hop path questions (WC-2H), and conjunctive questions (WC-C). WC-M is the mixture of WC-1H and WC-2H. 
The statistics of {\em WorldCup2014} are listed in Table~\ref{tab:dataset}.

\begin{table}[htbp]
	\centering
	\resizebox{!}{0.75cm}
	{
		\begin{tabular}{|c I c I c I c| c |}
			\hline
			Dataset & \#Entity & \#Relation   & \#Question &Exemplar Question\\
			\hline
			PQ-2H / 3H & 2,215 &14  &1,908 / 5,198 & What does the son of princess\_Sophia's mom do for a living? \\
			PQL-2H / 3H & 5,035 & 364  &1,594 / 1,031 & What is the notable type of Jody\_Harris's profession? \\
			WC-1H / 2H / M / C & 1,127 & 6 & 6,482 / 1,472 / 7,954 / 2,208 & Name a player who plays at Forward position from Mexico?\\
			\hline
		\end{tabular}
	}
	\caption{\label{tab:dataset} \small Statistics and exemplar questions of PathQuestion (PQ), PathQuestion-Large (PQL) and WorldCup2014 (WC). }
\end{table}

\section{Experiment and Evaluation}
\subsection{Implementation Details}
ADAM optimizer~\cite{Adam} was used for parameter optimization. 
The dimension of all the embeddings (words in question, entities and relations in KB) was set as $d_x=d_e=d_r=50$. 
The hyper-parameter $\lambda$ (see Eq.~\ref{eq:loss}) is set to $1$ . 
We partitioned the entire dataset into the train/valid/test subset with a proportion of $8:1:1$ and set the batch size as $50$.

\subsection{Performance of Question Answering}
In this section, we evaluated the performance of multi-relation question answering on {\bf PathQuestion} and {\bf WorldCup2014} respectively. To further show that IRN is able to handle more challenging datasets, we evaluated the model with two configurations:\\
%
 {\bf Incomplete KB } ~To simulate the real KBs which are often far from complete, we removed half of the triples (entities and relations are retained but the connections between entities were removed) from the KB of the PQ-2H dataset.\\
 {\bf Unseen KB} ~To simulate a real QA scenario where out-of-vocabulary(OOV) words is one of the major challenges, we removed questions whose answer path includes relation {\em ``Cause\_of\_Death"}, {\em ``Gender"} or {\em ``Profession"} from the PQ-2H training set. The models need to cope with questions related to these three OOV relations during the test.

Several baselines
are included here: \\
{\bf Embed}~\cite{EmbedQA} deals with factoid QA over KB by matching a question with an answer in the embedding spaces.\\
{\bf Subgraph}~\cite{SubgraphEmbed} improves the {\em Embed} model by enriching the representation of an answer entity with the entity's subgraph.\\
{\bf Seq2Seq}~\cite{Seq2Seq} is a simplified seq2seq semantic parsing model, which adopts an LSTM to encode the input question sequence and another LSTM to decode the answer path.\\
{\bf MemN2N}~\cite{MemN2N} is an end-to-end memory network that can be used for reading comprehension and question answering. 
The memory units consist of the related triples in a local subgraph of the corresponding answer path, 
where the settings are the same as ~\cite{SimpleQuestion}.\\
{\bf KVMemN2N}~\cite{KVMemN2N} improves the MemN2N for KBQA as it divides the memory into two parts: the key memory stores the head entity and relation while the value memory stores the tail entity.\\
{\bf IRN-weak} is our model that employs only supervision from the final answer entity rather than the complete answer path.
This can be implemented by simply ignoring the loss from the intermediate hops except the final entity in Eq.~\ref{eq:loss}.

The performance is measured by accuracy\footnote{Reported accuracy is the average accuracy of five repeated runs.}: correct if a predicted entity is in the answer set of input question. Since there are many 1-to-many relations in Freebase and WC2014, a question may have several possible answer paths, resulting in multiple answers.
For example, given the question {\em ``How old is Obama's daughter?"}, the original path can be {\em ``Barack\_Obama $\xrightarrow{Children}$ Malia\_Obama $\xrightarrow{Age}$18"} or {\em ``Barack\_Obama $\xrightarrow{Children}$Sasha\_Obama $\xrightarrow{Age}$ 14"}, thus the answer can be either {\em ``18"} or {\em ``14"}. For this question, either answer is correct.

The results in Table~\ref{tab:accuracy} demonstrate that our system outperforms the baselines on single-relation questions (WC-1H), 2-hop-relation questions (PQ-2H/PQL-2H/WC-2H) as well as 3-hop-relation questions (PQ-3H/PQL-3H). 
Furthermore, assembled IRNs obtain strong performance when dealing with conjunctive questions in WC-C .

We have further observations as follows:\\
	$\bullet$ {\em IRN-weak} outperforms {\em Embed} and {\em Subgraph}, indicating that multi-hop reasoning indeed helps to answer complex questions even when our model is trained end-to-end in the same configuration of weak supervision\footnote{Only supervision from question-answer pairs, but without answer path information from KB.}.\\ 
	$\bullet$ The {\em Seq2Seq} baseline performs worse than {\em IRN}. Though they are both interpretable, {\em IRN} is more powerful when dealing with complicated KBs and questions.\\
	$\bullet$ {\em IRN} is better than {\em MemN2N} and {\em KVMemN2N} on most of the datasets, 
	and both models are much better than other baselines using the path information. Note that the memory in {\em (KV)MemN2N} consists of fact triples which are distilled from KB using answer path. 
	In this sense, {\em (KV)MemN2N} indirectly employs strong supervision from answer path. 
	In contrast, {\em IRN} has a better (or easier) mechanism to supervise the reasoning process thanks to its interpretable framework.\\
	$\bullet$ The highest accuracy on PQL-2H/3H reveals that {\em IRN} performs better when faced with larger datasets. IRN deals with relations and entities separately, where the number of entities and relations is much less than that of triples. However, {\em (KV)MemN2N} has to handle much more triples in its memory.\\
	$\bullet$ {\em IRN} is more robust than the baselines ($\downarrow 3.7\%$ vs. $\downarrow 2.3\%$ ) when dealing with incomplete KB, which is probably because auxiliary KB embedding training facilitates the prediction of missing triples. While the baselines are more sensitive to the incomplete information stored in the memory units.\\
	$\bullet$ Both {\em IRN} and the baselines degrade remarkably (0.9$\rightarrow$0.5) in the unseen setting because wrong distributed representations are influential in embedding-based QA models. 
	In addition, the size of the training set is much smaller than that of the original PQ-2H, which also leads to worse performance.\\
	$\bullet$ {\em IRN} is more interpretable compared with {\em (KV)MemN2N },  
	attributed to the structure of {\em IRN}. The relation/entity predicted at each hop is a part of the answer path.
	The intermediate outputs offer the possibility to trace the complete reasoning process, diagnose failures, and manipulate answer prediction through intermediate interactions (see Section~\ref{sect:interpretable_exp}).

\begin{table*}
	\centering
	\small
	\setlength{\tabcolsep}{4pt}
	\begin{tabular}{|c| cIc | cIc | cIcIcIc| cIc |}
		\hline
		&\multicolumn{2}{c|}{PathQuestion}&\multicolumn{2}{c|}{PathQuestion Large} &\multicolumn{4}{c|}{WorldCup2014} &\multicolumn{2}{c|}{Challenging PQ-2H}\\
		\cline{2-11}
		&PQ-2H &PQ-3H & PQL-2H & PQL-3H & WC-1H &WC-2H &WC-M &WC-C &Incomplete &Unseen   \\
		\hline
		Random  & 0.151    & 0.104 & 0.021 & 0.015 & 0.085 & 0.064 & 0.053 & 0.073 & - & -  \\
		Embed & 0.787    & 0.483 & 0.425 & 0.225 & 0.448 & 0.588 & 0.518 & 0.642 & - & -\\
		Subgraph  & 0.744  & 0.506 & 0.500 & 0.213  & 0.448 & 0.507 & 0.513 & 0.692 & - & - \\
		IRN-weak({\em Ours})  & 0.919   & 0.833 & 0.630 & 0.618 & 0.749 & 0.921 & 0.786 & 0.837  & - & - \\
		\hline
		Seq2Seq & 0.899 & 0.770   &  0.719 & 0.647  & 0.537 & 0.548  &0.538 & 0.577 & - & -  \\
		MemN2N  & 0.930   &  0.845 & 0.690 & 0.617 & 0.854 & 0.915 & \bf 0.907 & 0.733  & 0.899($\downarrow 3.3\% $) & 0.558 \\
		KVMemN2N  & 0.937  &  \bf 0.879 & 0.722 & 0.674  &\bf 0.870 & 0.928 & 0.905 & 0.788 & 0.902($\downarrow 3.7\% $) & 0.554 \\
		IRN({\em Ours})  &\bf 0.960   &  0.877 & \bf 0.725 & \bf 0.710 & 0.843 & \bf 0.981 &\bf 0.907 & \bf 0.910  &  0.937($\downarrow 2.3\% $) & 0.550 \\
		\hline
	\end{tabular}
	\caption{\label{tab:accuracy}\small  Accuracy on different QA datasets. WC-C is for conjunctive questions while other datasets for path questions. Challenging PQ-2H are two more difficult configurations of PQ-2H. The models in the second block utilize the answer path information but those in the first block do not.} 
\end{table*}

\subsection{Interpretable Path Reasoning}
\label{sect:interpretable_exp}
In this section, we demonstrated how IRN is interpretable by both quantitative and qualitative analysis.
For the quantitative analysis, we can measure how it performs during the reasoning process by investigating the prediction accuracy of intermediate relations and entities. 
In this task, 
we collected all the relations and entities with largest probabilities (see Eq.~\ref{eq:g} and Eq.~\ref{eq:o}) at each hop  $\{r^1, a^1, r^2,..,a^H\}$ and compared these intermediate outputs with the ground truth $\{r_1, e_1, r_2, ..,a\}$.  As for KVMemN2N, we also fetched an entity at each hop by an answer distribution, similar to that at the final hop. 
\begin{table}[htbp]
	\centering
	\small
	\resizebox{!}{1.3cm}
	{
	\begin{tabular}{|c| cIc | cIc | cIc | cIc | cIc  | cIc |}
		\hline
		& \multicolumn{2}{c|}{$r_1$} & \multicolumn{2}{c|}{$e_1$} & \multicolumn{2}{c|}{$r_2$}  & \multicolumn{2}{c|}{$e_2$} & \multicolumn{2}{c|}{$r_3$} & \multicolumn{2}{c|}{$a$} \\
		\cline{2-13}
		& IRN & KVM & IRN & KVM& IRN & KVM& IRN & KVM& IRN & KVM& IRN & KVM\\
		\hline
		PQ-2H & 1.000 & NA & 0.957 &0.016 & 1.000 & NA&NA&{NA}&NA&NA& 0.934 & 0.916 \\
		PQL-2H & 0.968 & NA & 0.722 & 0.083 & 0.836 & NA&NA&{NA}&NA&NA&0.673&0.676 \\
		WC-2H & 1.000 & NA & 0.531 & 0.000 & 1.000 & NA&NA&{NA}&NA&NA&0.528& 0.382\\
		PQ-3H & 1.000 & NA & 0.883 & 0.003 & 1.000 & NA&0.772 & 0.001 &1.000 & NA &0.738&0.774 \\
		PQL-3H & 0.808 & NA & 0.721 & 0.019 & 0.702 & NA & 0.721 & 0.007 &0.683 & NA & 0.608 &0.600 \\
		\hline
	\end{tabular}
	}
	\caption{\label{tab:path-result}\small Accuracy at different hops along the answer path from IRN and KVMemN2N (KVM).
		$r_i$ indicates the relation at hop $i$, $e_i$ indicates the entity at hop $i$. $a$ indicates the final answer.
		NA means not applicable.
	}
\end{table}

According to the structure of IRN, the relation/entity predicted at each hop constitutes an answer path.
Results\footnote{Note here that only if an output matches the labeled entity exactly, the prediction will be judged as correct.Thus, the accuracy here has a different definition from that in Table~\ref{tab:accuracy}.} in Table~\ref{tab:path-result} indicate that IRN can predict intermediate entities more accurately than final answers, due to the cascading errors in the consecutive prediction. 

Though KVMemN2N (KVM) predicts the exact answers well, it lacks interpretability. On the one hand, KVM can not predict relations to trace the answer path. On the other hand, the hops in KVM all aim at finding the answer entity rather than the intermediate entities along the answer path. 

\begin{figure*}[htbp]
  \centering
  \small
  \includegraphics[width=0.95\linewidth]{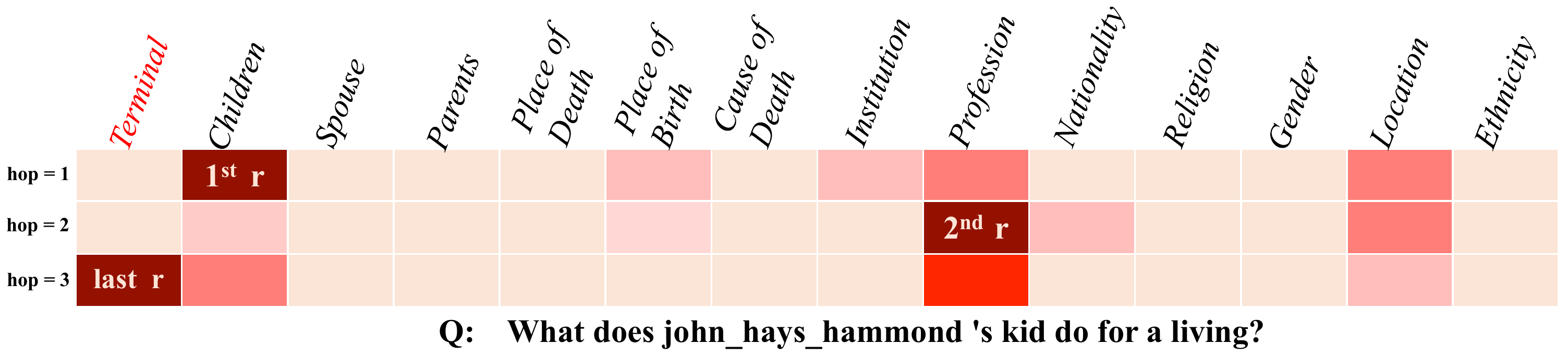}
  \caption{\small The predicted relations at each hop. Each row represents a probability distribution over relations. Darker color indicates larger probability. The terminal relation is highlighted in red.}
  \label{fig:gate}
\end{figure*}

To illustrate how our model parses a question and predicts relations hop-by-hop, we studied the distributions over all the relations ($\boldsymbol{g}^h$, see Eq.~\ref{eq:g}) and chose an example from PathQuestion as shown in Figure~\ref{fig:gate}. 
It is clear that IRN is able to derive the relations in correct order. For question {\em ``What does john\_hays\_hammond's kid do for a living?"}, IRN first detects relation {\em Children} (the corresponding word/phrase in the question is {\em kid}) and then {\em Profession} ({\em what does..do}). 
When detecting the {\em Terminal} relation, IRN will stop the analysis process. 

Our model can map relations in KB to words in question. We sampled some relations in KB and projected them to the question space by $\boldsymbol{r}^q = \boldsymbol{M}_{rq}\boldsymbol{r}$ (see Eq.~\ref{eq:Mrq}). We then obtained words whose embeddings are most close to $\boldsymbol{r}^q$ measured by cosine similarity $cos(\boldsymbol{r}^q,\boldsymbol{x}_i)$. The result in Table~\ref{tab:relation2word} indicates that IRN can establish reasonable mapping between KB relation and natural language, such as linking {\em Profession } to words like {\em working, profession}, and {\em occupation }. 
Besides mapping to single words, relation in KB can be associated with some complicate templates, such as {\small {\em Profession } $\rightarrow$ {\em ``what does ... do" }}. 
\begin{table*}[htbp]
\centering
\small
\begin{tabular}{|c | c |}
  \hline
  Relation & Similar words in natural language questions \\
  \hline
  Profession &  profession, do, working, occupation\\
  Insititution &  institution, organization, work, where\\
  Religion & faith, religion, what, belief\\
  Cause\_of\_Death & died, killed, how, death\\
  Place\_of\_Birth &  hometown, born, city, birthplace\\
  \hline
\end{tabular}
\caption{\label{tab:relation2word}\small  Most similar words in questions for some exemplar relations.}
\end{table*}

The above analysis demonstrates that our model is interpretable. Specifically, IRN has merits at:
\\
	$\bullet$Providing a traceable reasoning path for question answering. With the aid of these intermediate entities and relations, we can obtain not only the final answer but also the complete path that infers the answer.
	\\
	\\
	$\bullet$Facilitating failure diagnosis. For instance, 
	IRN fails to answer the question {\small \em ``Where did the child of Joseph\_P\_Kennedy\_Sr die ?"}. The true answer path should be {\small \em ``Joseph\_P\_Kennedy\_Sr $\xrightarrow{Children}$Patricia\_kennedy\_Lawford $\xrightarrow{Place\_of\_Death}$New\_York\_County"}. However, the middle entity decided by IRN is {\small \em ``Rosemary\_Kennedy"} who is also a child of {\small \em ``Joseph\_P\_Kennedy\_Sr"}, but her death is not included in KB.
	\\
	\\
	$\bullet$Allowing manual manipulation in answer prediction.
	 We updated the state (Eq.~\ref{eq:state}) and the question (Eq.~\ref{eq:Mrq}) in IRN with the ground-truth relation vectors and compared the performance. The higher accuracy in Table~\ref{tab:correct} implies that we can improve the final prediction by correcting intermediate predictions.
\begin{table}[htbp]
	\centering
	\small
	\begin{tabular}{|c|c I c I c I c|}
		\hline
		Dataset & PQ-2H & PQ-3H & PQL-2H & PQL-3H \\
		\hline
		Acc & 0.980 ($\uparrow 2.0\% $) & 0.900 ($\uparrow 2.3\% $) & 0.755 ($\uparrow 3.0\% $) & 0.744 ($\uparrow 3.4\% $)\\
		\hline
	\end{tabular}
	\caption{\label{tab:correct}\small Accuracy when the intermediate predictions are replaced by ground truth.
	}
\end{table}

\section{Conclusion}
\label{sect:conclusion}
We present a novel Interpretable Reasoning Network which is able to make reasoning hop-by-hop and then answer multi-relation questions.
Our model is interpretable in that the intermediate predictions of entities and relations are traceable 
and the complete reasoning path is observable. This property enables our model to facilitate reasoning analysis, failure diagnosis, and manual manipulation in answer prediction.
Results on two QA datasets demonstrate the effectiveness of the model on multi-relation question answering.

As future work, there is much room for complex question answering. For instance, answering { \em ``How old is Obama's younger daughter?"} needs to handle arithmetic operation. Furthermore, multi-constraint questions will also be considered in this framework.

%

\section*{Acknowledgements}
This work was partly supported by the National Science Foundation of China under grant No.61272227/61332007 and the National Basic Research Program (973 Program) under grant No. 2013CB329403.

\bibliographystyle{acl}
\bibliography{reference}

\hfill
\section*{Appendix A. PathQuestion Construction and Question Templates}
\label{sec:supplemental}
We constructed a synthesis dataset by generating questions with templates. The knowledge base for PathQuestion has more than 60,000 triples which are adopted from FB13~\cite{FB13} with 13 relations and thousands of entities. As for PathQuestion-Large, we adopted another more complex subset of Freebase~\cite{Bollacker2008Freebase}. 
\textbf{First}, we extracted all the paths with two hops ($<e_s,r_1,e_1,r_2,a>$), or three hops ($<e_s,r_1,e_1,r_2,e_2,r_3,a>$) among these triples. \textbf{Second}, we crafted templates to generate natural language questions from these paths. \textbf{Last}, we collected question and answer path pairs $(q,<e_s,r_1,e_1,...,a>)$ to construct the {\em PathQuestion (PQ)} dataset.

We crafted templates to transfer an answer path extracted from KB 
to natural language questions.
To make the generated questions analogical to real-world questions, those templates are firstly written manually, and then enriched by replacing synonyms. Besides, we searched for different syntactical structures and paraphrases in real-world datasets including WebQuestions~\cite{WebQuestions} and WikiAnswers~\cite{Fader2013Paraphrase} as well as on the Internet. In this manner, the templates have been greatly diversified and are much closer to real questions.

Synonyms used in templates for PathQuestion are shown in Table~\ref{tab:synonyms} and templates for 2-hop paths (PQ-2H) are shown in Table~\ref{tab:templates}.
The datasets are available at~\url{https://github.com/zmtkeke/IRN}.
\begin{table}[htbp]
	\small
	\centering
	\begin{tabular}{|l|l|}
		\hline 
		\bf Relation & \bf Synonyms \\ 
		\hline
		\multirow{2}{2cm}{Spouse} & couple, wife, husband\\
		\cline{2-2} 
		& other half, darling\\
		\hline
		\multirow{2}{2cm}{Children}& child, offspring, kid\\
		\cline{2-2}
		& daughter, son, heir\\
		\hline
		\multirow{2}{2cm}{Parents}& parent, father, mother\\
		\cline{2-2}
		& dad, mom\\
		\hline
		\multirow{1}{1cm}{Profession} & job, occupation, work\\
		\hline
		\multirow{2}{2cm}{Institution}  & organization\\
		\cline{2-2}
		& educational institution\\
		\hline
		\multirow{1}{1cm}{Ethnicity}  & race\\
		\hline
		\multirow{1}{1cm}{Gender}  & sex\\
		\hline
		\multirow{1}{1cm}{Nationality}  & nation, country\\
		\hline
		\multirow{1}{1cm}{Location}  & address\\
		\hline
		\multirow{2}{2cm}{Religion}  & faith, religious belief\\
		\cline{2-2}
		& type of religion\\
		\hline
	\end{tabular}
	\caption{Natural language synonyms for relations appearing in questions in PathQuestion.}
	\label{tab:synonyms}
\end{table}

\begin{table}[htbp]
	\centering
	\small
	\begin{tabular}{|l|l|}
		\hline 
		\bf Path Pattern & \bf Question-templates \\ 
		\hline
		\multirow{6}{6cm}{Universal} & ``What is the $r_2$ of $e_s$'s $r_1$?"\\
		\cline{2-2} 
		& ``What is the $e_s$'s $r_1$'s $r_2$?"\\
		\cline{2-2} 
		& ``What is the $r_2$ of $r_1$ of $e_s$?"\\
		\cline{2-2} 
		& ``The $r_2$ of $e_s$'s $r_1$?"\\
		\cline{2-2} 
		& ``The $e_s$'s $r_1$'s $r_2$?"\\
		\cline{2-2} 
		& ``The $r_2$ of $r_1$ of $e_s$?"\\
		\hline
		Ask about a person & ``Who is the $r_2$ of $e_s$'s $r_1$?"\\
		\cline{2-2} 
		& ``What is the name of the $r_2$ of $e_s$'s $r_1$?"\\
		\hline
		\multirow{1}{2cm}{$r_2$=$r_1$=Parents/} & ``Who is the grand-$r_1$ of $e_s$?"\\
		\cline{2-2} 
		$r_2$=$r_1$=Children& ``What is the name of the grand-$r_1$ of $e_s$?"\\
		\hline
		\multirow{3}{3cm}{$r_2$=Ethnicity} & ``What $r_2$ is $e_s$'s $r_1$"\\
		\cline{2-2}
		& ``What is $e_s$'s $r_1$'s $r_2$ like?"\\
		\cline{2-2} 
		& ``What is $e_s$'s $r_1$'s $r_2$ about?"\\
		\hline
		\multirow{3}{3cm}{$r_2$=Institution}& ``Where does $e_s$'s $r_1$ work?"\\
		\cline{2-2}
		& ``Where does $e_s$'s $r_1$ work for?"\\
		\cline{2-2}
		& ``Which $r_2$ does $e_s$'s $r_1$ work for?"\\
		\hline
		\multirow{2}{2cm}{$r_2$=Nationality}& ``Which nationality is $e_s$'s $r_1$?"\\
		\cline{2-2}
		& ``Where does $e_s$'s $r_1$ come from?"\\
		\hline
		\multirow{4}{4cm}{$r_2$=Religion} & ``What $r_2$ does $e_s$'s $r_1$ follow?"\\
		\cline{2-2}
		& ``What $r_2$ is $e_s$'s $r_1$?"\\
		\cline{2-2}
		& ``What $r_2$ does $e_s$'s $r_1$ have?"\\
		\cline{2-2}
		& ``What $r_2$ is $e_s$'s $r_1$ practice?"\\
		\hline
		\multirow{2}{2cm}{$r_2$=Gender}  & ``What $r_2$ is $e_s$'s $r_1$?"\\
		\cline{2-2}
		& ``Is $e_s$'s $r_1$ a man or a woman?"\\
		\hline
		\multirow{3}{3cm}{$r_2$=Location}& ``Where is $e_s$'s $r_1$ living?"\\
		\cline{2-2}
		& ``Where is $e_s$'s $r_1$ staying?"\\
		\cline{2-2}
		& ``Please tell me $e_s$'s $r_1$ present address."\\
		\hline
		\multirow{5}{5cm}{$r_2$=Profession} & ``What does $e_s$'s $r_1$ do?"\\
		\cline{2-2}
		& ``What is $e_s$'s $r_1$ working on?"\\
		\cline{2-2}
		& "What is $e_s$'s $r_1$?"\\
		\cline{2-2}
		& ``What line of business is $e_s$'s $r_1$ in?"\\
		\cline{2-2}
		& ``What does $e_s$'s $r_1$ do for a living?"\\
		\hline
		\multirow{7}{7cm}{$r_2$=Cause\_of\_Death} & ``Why $e_s$'s $r_1$ died?"\\
		\cline{2-2}
		& ``How $e_s$ died?"\\
		\cline{2-2}
		& ``What's the reason of $e_s$'s $r_1$'s death?"\\
		\cline{2-2}
		& ``What caused the death of $e_s$'s $r_1$?"\\
		\cline{2-2}
		& "What killed the $e_s$'s $r_1$?"\\
		\cline{2-2}
		& ``What made the $e_s$'s $r_1$ dead?"\\
		\cline{2-2}
		& ``What did $e_s$'s $r_1$ die from?"\\
		\hline
		\multirow{3}{3cm}{$r_2$=Place\_of\_Death} & ``Where did $e_s$'s $r_1$ die?"\\
		\cline{2-2}
		& ``Where did the $r_1$ of $e_s$ die?"\\
		\cline{2-2}
		& ``What city did $e_s$'s $r_1$ die?"\\
		\hline
		\multirow{4}{4cm}{$r_2$=Place\_of\_Birth} & ``Where did $e_s$'s $r_1$ born?"\\
		\cline{2-2}
		& ``What city did $e_s$'s $r_1$ born?"\\
		\cline{2-2}
		& ``What is the hometown of $e_s$'s $r_1$?"\\
		\cline{2-2}
		& ``What is $e_s$'s $r_1$'s birthplace?"\\
		\hline
	\end{tabular}
	\caption{Templates for generating natural language questions from answer paths in PQ-2H.}
	\label{tab:templates}
\end{table}

\end{document}